\documentclass[10pt,twocolumn,letterpaper]{article}

\usepackage{cvpr}              

\usepackage{tabularx}   
\usepackage{booktabs}   
\usepackage{makecell}

\usepackage{multirow}

\definecolor{cvprblue}{rgb}{0.21,0.49,0.74}
\usepackage[pagebackref,breaklinks,colorlinks,allcolors=cvprblue]{hyperref}

\def\paperID{16168} 
\def\confName{CVPR}
\def\confYear{2026}

\title{PAVE: An End-to-End Dataset for Production Autonomous Vehicle Evaluation}

\author{
Xiangyu Li \and 
Chen Wang \and 
Yumao Liu \and 
Dengbo He \and 
Jiahao Zhang \and 
Ke Ma \textsuperscript{*}\\
The Hong Kong University of Science and Technology (Guangzhou) \\
{\tt\small xiangyuli2@hkust-gz.edu.cn, kema@hkust-gz.edu.cn}\\
}

\begin{document}
\maketitle
\begin{abstract}

Most existing autonomous-driving datasets (\eg KITTI, nuScenes, and the Waymo Perception Dataset), collected by human-driving mode or unidentified driving mode, can only serve as early training for the perception and prediction of autonomous vehicles (AVs). To evaluate the real behavioral safety of AVs controlled in the black box, we present the first end-to-end benchmark dataset collected entirely by autonomous-driving mode in the real world. This dataset contains over 100 hours of naturalistic data from multiple production autonomous-driving vehicle models in the market. We segment the original data into 32,727 key frames, each consisting of four synchronized camera images and high-precision GNSS/IMU data (0.8 cm localization accuracy). For each key frame, 20 Hz vehicle trajectories spanning the past 6 s and future 5 s are provided, along with detailed 2D annotations of surrounding vehicles, pedestrians, traffic lights, and traffic signs. These key frames have rich scenario-level attributes, including driver intent, area type (covering highways, urban roads, and residential areas), lighting (day, night, or dusk), weather (clear or rain), road surface (paved or unpaved), traffic and vulnerable road users (VRU) density, traffic lights, and traffic signs (warning, prohibition, and indication). To evaluate the safety of AVs, we employ an end-to-end motion planning model that predicts vehicle trajectories with an Average Displacement Error (ADE) of 1.4 m on autonomous-driving frames. The dataset continues to expand by over 10 hours of new data weekly, thereby providing a sustainable foundation for research on AV driving behavior analysis and safety evaluation. The PAVE dataset is publicly available at \url{https://hkustgz-my.sharepoint.com/:f:/g/personal/kema_hkust-gz_edu_cn/IgDXyoHKfdGnSZ3JbbidjduMAXxs-Z3NXzm005A_Ix9tr0Q?e=9HReCu}.


\end{abstract}

\section{Introduction}
\label{sec:intro}

Autonomous vehicles (AVs) have long promised to enhance road safety, but this goal remains unfulfilled. For human drivers in the United States, the crash probability is roughly $2.6 \times 10^{-5}$ per mile based on the SHRP2 Naturalistic Driving Study~\cite{dingus2016driver}, or about $1.9 \times 10^{-6}$ per mile from NHTSA police-reported data~\cite{NHTSA2023}. In contrast, state-of-the-art AVs operating in California report disengagement rates of $6.54 \times 10^{-4}$ per mile in 2021~\cite{CAReport2021} and $4.53 \times 10^{-4}$ in 2024~\cite{CAReport2024}, showing steady improvement but still remaining far above human crash levels. This gap highlights that, despite rapid progress, AVs still fall short of human-level safety, mainly because they struggle with long-tail scenarios—rare and unpredictable situations~\cite{makansi2021exposing} (such as adverse weather, unusual road users, or complex interactions). The challenge is further amplified in end-to-end systems, where decisions come from black-box models~\cite{omeiza2021explanations} with limited interpretability. This opacity makes it especially difficult to evaluate the behavior and safety of AVs in complex traffic scenarios—one of the central goals of AV research. 

Achieving this requires large-scale end-to-end datasets that jointly capture both perception inputs (i.e., raw sensor data) and trajectory outputs (i.e., continuous positions/velocities). Here, perception inputs refer to key-frame raw sensor data, such as camera or LiDAR measurements, while trajectory outputs refer to positions/velocities sequences longer than 9 seconds and sampled at a frequency of at least 2\,Hz. Based on the definition of perception and trajectory data, we broadly categorize existing autonomous driving datasets into \emph{perception}, \emph{motion}, and \emph{end-to-end} datasets. The perception dataset provides raw sensor data but lacks continuous trajectory outputs. A trajectory dataset contains continuous trajectory outputs but does not include the corresponding perception inputs. In contrast, an end-to-end dataset provides both perception and trajectory data. Yet the released data contain only limited autonomous driving segments that are blended with human-driving data, preventing any clear identification between the two modes. To the best of our knowledge, no existing datasets provide an identified autonomous driving mode and sufficiently rich AV data with aligned perception and motion information.






Thus, we present the Production Autonomous Vehicle Evaluation (PAVE) dataset, a large-scale dataset that integrates synchronized multi-camera perception and trajectory data collected under identified human and autonomous driving modes. The PAVE dataset provides wide visual coverage from multiple cameras and centimeter-level localization accuracy through GNSS/IMU. In total, it contains over 140 hours of real-world operation across various environments and vehicle brands/models. The PAVE dataset is organized into 32727 key frames, selected as representative frames of driving scenarios. These key frames support a variety of perception tasks such as detection, tracking, and motion planning, and enable detailed analysis of autonomous driving behavior and safety evaluation under diverse traffic scenarios.

The main contributions of our work are summarized as follows:
\begin{itemize}
    \item We present a large-scale end-to-end dataset that combines perception and trajectory data identified for autonomous driving.
    \item     The PAVE dataset enables inverse modeling of the production AV models from perception and trajectory data.
    \item The PAVE dataset enables interpretable driving behavior analysis and safety evaluation under diverse traffic scenarios across different vehicle brands/models.

\end{itemize}


\section{Related Work}
\label{sec:related}



In this work, we use the term \emph{autonomous driving} - and the corresponding vehicles, AVs - to denote all driver-assistance modes, where the system performs both longitudinal and lateral control. Although the driver may still be expected to monitor the environment or take over upon request (as in typical Level 2/Level 2+/Level 3-like systems), the AV executes the primary dynamic driving tasks during operation. Typical examples include Tesla FSD, XPeng NOA, and Huawei ADS/NCA.


\paragraph{Perception datasets.}

Early perception benchmarks such as KITTI~\cite{geiger2013vision} and the Waymo Perception dataset~\cite{sun2020scalability} provide synchronized camera--LiDAR data with high-quality 3D annotations. KITTI is recorded entirely under human driving. Waymo covers multiple U.S.\ cities, but the released labels do not identify the driving mode. The ZOD Frame~\cite{alibeigi2023zenseact} adds high-resolution multimodal data with detailed 2D/3D annotations. These datasets offer rich sensor diversity for detection and tracking. However, they do not show autonomous-driving behavior or provide identifiable driving-mode information.

\paragraph{Motion datasets.}

Beyond perception, several datasets focus on motion prediction. The Waymo Motion Dataset~\cite{ettinger2021large} contains roughly 570 hours of trajectories and over 100k annotated scenes, yet it does not identify which segments correspond to AV versus HV. The OpenPAV Dataset contains 13 open-source datasets from 7 providers of multiple AVs, but it provides only trajectory data without synchronized perception data.

\paragraph{End-to-end datasets.}

End-to-end datasets combine perception and trajectory data, providing a unified view of sensing and motion. nuScenes~\cite{caesar2020nuscenes} offers multimodal 20-second sequences with 3D annotations. The Waymo E2E dataset~\cite{waymo_e2e_dataset} integrates eight-camera perception with aligned trajectories across over 4,000 scenes, and ZOD Sequences~\cite{alibeigi2023zenseact} provide 20-second segments with camera, LiDAR, and high-precision GNSS/IMU. However, although portions of these data include autonomous driving mode, the released annotations do not identify whether segments were recorded under human or autonomous driving modes. Consequently, these datasets do not expose identifiable autonomous-driving behavior. 

A detailed comparison of representative datasets is shown in \cref{tab:dataset_comparison}.

\begin{table*}[t]
  \caption{
  Comparison of representative autonomous driving datasets across perception, motion, and end-to-end categories.
  Abbreviations: 
  Human = human driving, 
  Auto = autonomous driving, 
  H\,+\,A (labeled) = human + autonomous data with explicit driving-mode labels, 
  H\,+\,A (unlabeled) = both Human and Auto without distinction. 
  RGB imgs = RGB images, Ann. frames = annotated frames, Cams = cameras, Accuracy = ego-localization accuracy, Attitude = pose availability, Ann. scenarios = annotated scenarios, 
  Avg. speed = average driving speed (m/s). 
  Tasks: Det. = detection, Seg. = segmentation, Track. = tracking, Motion Pred. = motion prediction,  Motion Plan. = motion planning, Eval. = safety evaluation. 
  Symbols: -- = not reported, N/A = not applicable, $\sim$ = approximately.
  }
  \label{tab:dataset_comparison}
  \centering
  \scriptsize
  \setlength{\tabcolsep}{3pt}
  \begin{tabularx}{\textwidth}{c|c|ccccccc|ccc|cc|cc}
    \toprule
    \textbf{Category} & \textbf{Dataset} 
      & \multicolumn{7}{c|}{\textbf{General}} 
      & \multicolumn{3}{c|}{\textbf{Perception}} 
      & \multicolumn{2}{c|}{\textbf{Trajectory}} 
      & \multicolumn{2}{c}{\textbf{Tasks}} \\
    \midrule
      &  & Year & Scenes & Size (h) & \makecell{Veh.\\models} & \makecell{Driving\\mode} & Locations & \makecell{Avg.\\speed} 
      & \makecell{RGB\\imgs} & \makecell{Ann.\\frames} & Cams 
      & Accuracy & Attitude 
      & \makecell{Ann.\\scenarios} & Tasks \\
    \midrule
    \multirow{3}{*}{Perception} 
     & KITTI        & 2012 & 22 & 1.5 & 1 & Human & Karlsruhe    
                 & 9.7 
                 & 15k & 15k & 4 
                 & 0.02 m & Yes 
                 & No & Det.\&Track. \\
     & \makecell{ZOD\\(Frame)} & 2023 & -- & -- & -- & \makecell{H\,+\,A\\(unlabeled)} & 14× Europe
                 & --
                 & 100k & 100k & 1 
                 & 0.01 m & Yes 
                 & No & Det., Seg. \\
     & \makecell{Waymo\\Perception}   & 2019 & 1k & 5.5 & 2 & \makecell{H\,+\,A\\(unlabeled)} & 3× USA       
                 & 9.76 
                 & 1M & 200k & 5 
                 & -- & -- 
                 & No & Det.\&Track. \\
     
    \midrule
    \multirow{2}{*}{Motion}
     & \makecell{Waymo\\Motion} & 2021 & 100k & 570 & 2 & \makecell{H\,+\,A\\(unlabeled)} & 6× USA
                 & 8.3
                 & N/A & N/A & N/A 
                 & -- & -- 
                 & No & Motion Pred. \\
     & OpenPAV      & 2024 & -- & -- & Multi & \makecell{H\,+\,A\\(unlabeled)} & Multi-source 
                 & 3.2-32.2 
                 & N/A & N/A & N/A 
                 & 0.01 - 0.1 m & -- 
                 & No & Motion Pred. \\
    \midrule
    \multirow{3}{*}{End-to-End}
     & nuScenes     & 2019 & 1k & 5.5 & 2 & \makecell{H\,+\,A\\(unlabeled)} & Boston, SG   
                 & 5.1 
                 & 1.4M & 40k & 6 
                 & $\leq$0.1 m & Yes 
                 & No & Det.\&Track. \\
     & \makecell{ZOD\\(Sequences)} & 2023 & 1473 & $\sim$8 & -- & \makecell{H\,+\,A\\(unlabeled)} & 14× Europe
                 & --
                 & 294k & 1473 & 1 
                 & 0.01 m & Yes 
                 & No & Det., Seg. \\
     & Waymo E2E    & 2025 & 4021 & $\sim$12 & 2 & \makecell{H\,+\,A\\(unlabeled)} & USA
                 & 5.8 
                 & $\sim$5.5M & -- & 8 
                 & -- & -- 
                 & No & Motion Plan. \\
     & \textbf{PAVE} & 2025 & 32727 & 140 & 5 & \makecell{\textbf{H\,+\,A}\\\textbf{(labeled)}} & \makecell{7 major cities\\in China \\and USA}
                 & 9.9
                 & 130k & 130k & 4
                 & \textbf{0.008 m} & \textbf{Yes} 
                 & \textbf{Yes} & \makecell{\textbf{Det., Eval.}\\\textbf{Motion Plan.}} \\
    \bottomrule
  \end{tabularx}
\end{table*}

\section{The Autonomous Driving Dataset}
\label{sec:dataset}

The PAVE dataset is a large-scale real-world driving dataset that includes both \emph{autonomous} and \emph{human} driving data, with each segment clearly identified by its driving mode. It is designed to support research in perception, motion planning, end-to-end learning, and safety evaluation of autonomous driving systems. The data were collected using multiple production vehicles equipped with \emph{autonomous driving systems}—including models such as the NIO~ET7, Tesla~Model~Y, XPeng~P7+, AITO~M7, etc.—each featuring a unified multi-sensor suite comprising four high-resolution cameras and a high-precision GNSS/IMU unit. Data collection was conducted across diverse environments in seven major cities in China and the USA, covering a broad range of road types and traffic scenarios. Each frame is annotated with comprehensive scenario attributes, including area type (\emph{highway, urban, residential}), lighting (\emph{day, dusk, night}), weather (\emph{clear, rain, snow}), road surface type (\emph{paved, unpaved}), vehicle density (\emph{high, mid, low}), vulnerable road user (VRU) density (\emph{high, mid, low}), presence of traffic lights, and traffic sign category (\emph{warning, prohibition, indication}). Each frame image includes 2D bounding box annotations for major traffic participants and objects, such as \emph{vehicles, pedestrians, motorcycles, traffic lights, and traffic signs}. Moreover, every frame is labeled with its driver mode (\emph{human or autonomous}) and the corresponding driver intent. In total, the PAVE dataset contains over 140 hours of real-world driving data, with synchronized multi-camera images and calibrated trajectory data. 


\subsection{Sensor Specifications}
\label{sec:sensor_spec}

\begin{figure}[t]
    \centering
    \includegraphics[width=\linewidth]{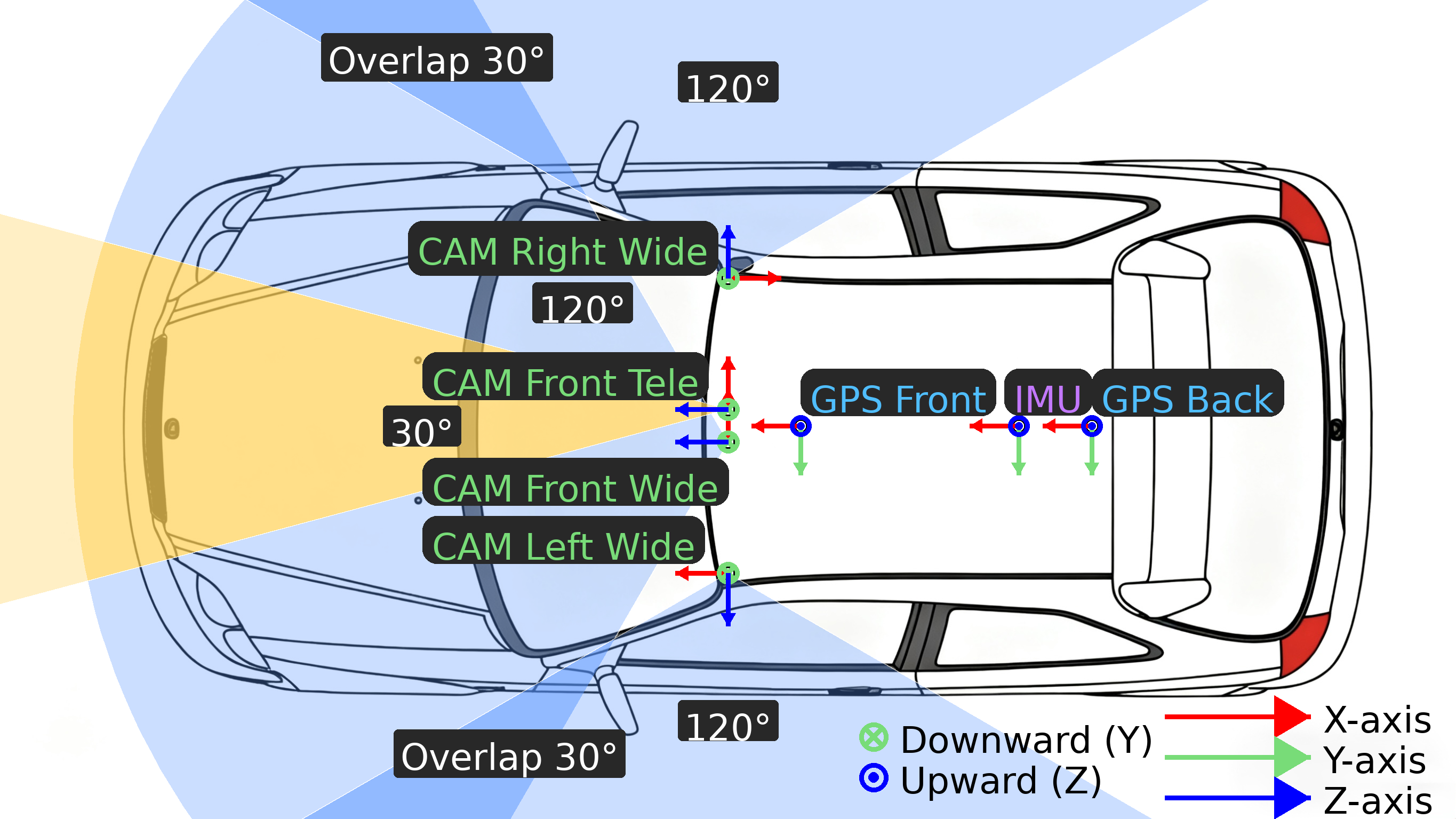}
    \caption{
        Sensor setup for our data collection platform.
    }
    \label{fig:sensor_layout}
\end{figure}

The data collection platform integrates a unified multi-sensor suite consisting of four high-resolution RGB cameras and a high-precision GNSS/IMU module. The cameras include one front-wide (horizontal field of view, FOV~$\sim$120°), one front-telephoto (FOV~$\sim$30°), and two side-wide cameras (left and right, each FOV~$\sim$120°), recording at a resolution of 2592×1944 and 30~FPS. The GNSS/IMU module performs integrated localization by fusing 20~Hz GNSS and 200~Hz IMU measurements, where GNSS localization is enhanced by Real-Time Kinematic (RTK) differential correction via the Continuously Operating Reference Station (CORS) network, achieving a localization accuracy of approximately 0.8~cm. The localization output provides latitude, longitude, altitude, velocity, heading, and the roll, pitch, and yaw angles derived from the IMU. All sensors are temporally synchronized using GNSS time as the global reference, ensuring precise alignment between video and trajectory data. As shown in \cref{fig:sensor_layout}, the sensors are mounted with fixed orientations, and the figure illustrates sensor layout and camera field-of-view coverage.

\subsection{Coordinate Systems}
\label{sec:coord_sys}

The PAVE dataset uses a unified coordinate framework, including global, vehicle, sensor, and image frames, as shown in \cref{fig:sensor_layout}.

\paragraph{Global Coordinate System.}
The global coordinate system is defined in the local East–North–Up (ENU) frame derived from the GNSS measurements. Localizations are represented in metric units (meters) relative to a fixed geographic reference point, allowing accurate spatial registration of all trajectories and annotations. All localization outputs are time-synchronized to this global ENU frame, which serves as the absolute reference for vehicle motion and sensor data alignment.

\paragraph{Vehicle Coordinate System.}
The vehicle coordinate system serves as the common reference frame for all sensors. The origin is defined at the center of the front bumper. The $X$-axis points forward along the vehicle’s heading direction, the $Y$-axis points to the left, and the $Z$-axis points upward, forming a right-handed coordinate system. All sensor extrinsic matrices are expressed relative to this vehicle frame.

\paragraph{Camera Coordinate System.}
Each camera coordinate system follows the pinhole model, with its origin located at the optical center of the lens. The $X$-axis points to the right in the image plane, the $Y$-axis points downward, and the $Z$-axis points forward along the optical axis. The extrinsic transformation matrix $\mathbf{T}_{\text{cam}\rightarrow\text{veh}}$ maps camera coordinates into the vehicle coordinate frame.

\paragraph{Image Coordinate System.}
The image coordinate system is a two-dimensional pixel frame used for visual annotations. The origin is located at the top-left corner of the image, with the $u$-axis increasing to the right and the $v$-axis increasing downward. All 2D bounding boxes and semantic labels are defined within this frame.

All transformation relationships among sensors are provided through extrinsic calibration files in the dataset, represented as $4\times4$ homogeneous matrices. These matrices enable direct coordinate transformation between each sensor frame, the vehicle frame, and the global reference frame, ensuring accurate spatial alignment across all modalities.

\subsection{Privacy Protection}
\label{sec:privacy}



All data in the PAVE dataset were carefully anonymized and processed in accordance with the privacy protection requirements of the People’s Republic of China, including the Personal Information Protection Law (PIPL) \cite{PIPL_2021_EN} and the Data Security Law (DSL) \cite{DSL_2021_EN}. To prevent the disclosure of personally identifiable information (PII), all human faces and vehicle license plates in the collected videos and images are automatically anonymized using an object-detection-based system developed by our team. A high-precision face and license-plate detection model is used to locate sensitive regions in each frame, after which these regions are blurred with a Gaussian filter to eliminate identifiable details. This processing is applied uniformly across all data modalities before storage or release, ensuring that no raw PII remains in the PAVE dataset.


\subsection{Data Categories}
\label{sec:data_cat}

The PAVE dataset is organized in the Frame category, designed for different research purposes and supports a wide range of perception, motion planning, and safety evaluation tasks.


\textbf{Frames:}  
Each frame in the PAVE dataset corresponds to an 11-second driving segment, and the released image is taken from the 6th second of that segment. The associated GNSS/IMU trajectory covers the past 6\, seconds and future 5\, seconds around the frame timestamp, providing vehicle motion and dynamic information at 20\,Hz. All trajectory points are expressed in the vehicle coordinate frame whose origin and orientation are defined by the vehicle pose at the 6th-second image frame. For each timestamp, the PAVE dataset provides the vehicle’s 3D position $(x, y, z)$ relative to this origin, the speed (scalar velocity magnitude), and the heading angle. In addition, the PAVE dataset provides the velocity and acceleration components in the $x$ (forward) and $y$ (lateral) directions of the vehicle coordinate frame. Each released frame includes four synchronized camera images (\textit{front\_wide}, \textit{front\_tele}, \textit{left\_wide}, and \textit{right\_wide}) together with the corresponding GNSS/IMU data. All images are automatically anonymized using our face and license-plate detection model (\cref{sec:privacy}). Each frame is comprehensively annotated with timestamp, 2D bounding boxes (for surrounding vehicles, pedestrians, motorcycles, traffic lights, and traffic signs), driving intent, driving mode, area\_type, lighting, weather, and road surface type, etc, as detailed in \cref{sec:gt_labels}. 


\subsection{Ground Truth Labels}
\label{sec:gt_labels}

We provide high-quality ground truth annotations for all camera and GNSS/IMU data. Each frame and sequence is manually labeled by professional annotators using industrial-grade labeling tools, followed by expert review to ensure high accuracy and consistency. Every frame contains precise timestamps that are globally synchronized across all sensors.

Each frame includes 2D bounding box annotations for major traffic participants and objects, including pedestrians, vehicles, motorcycles, traffic lights, and traffic signs. Each annotated instance is associated with a unique tracking ID to ensure object consistency across frames. 

\begin{figure}[t]
  \centering
  \includegraphics[width=\linewidth]{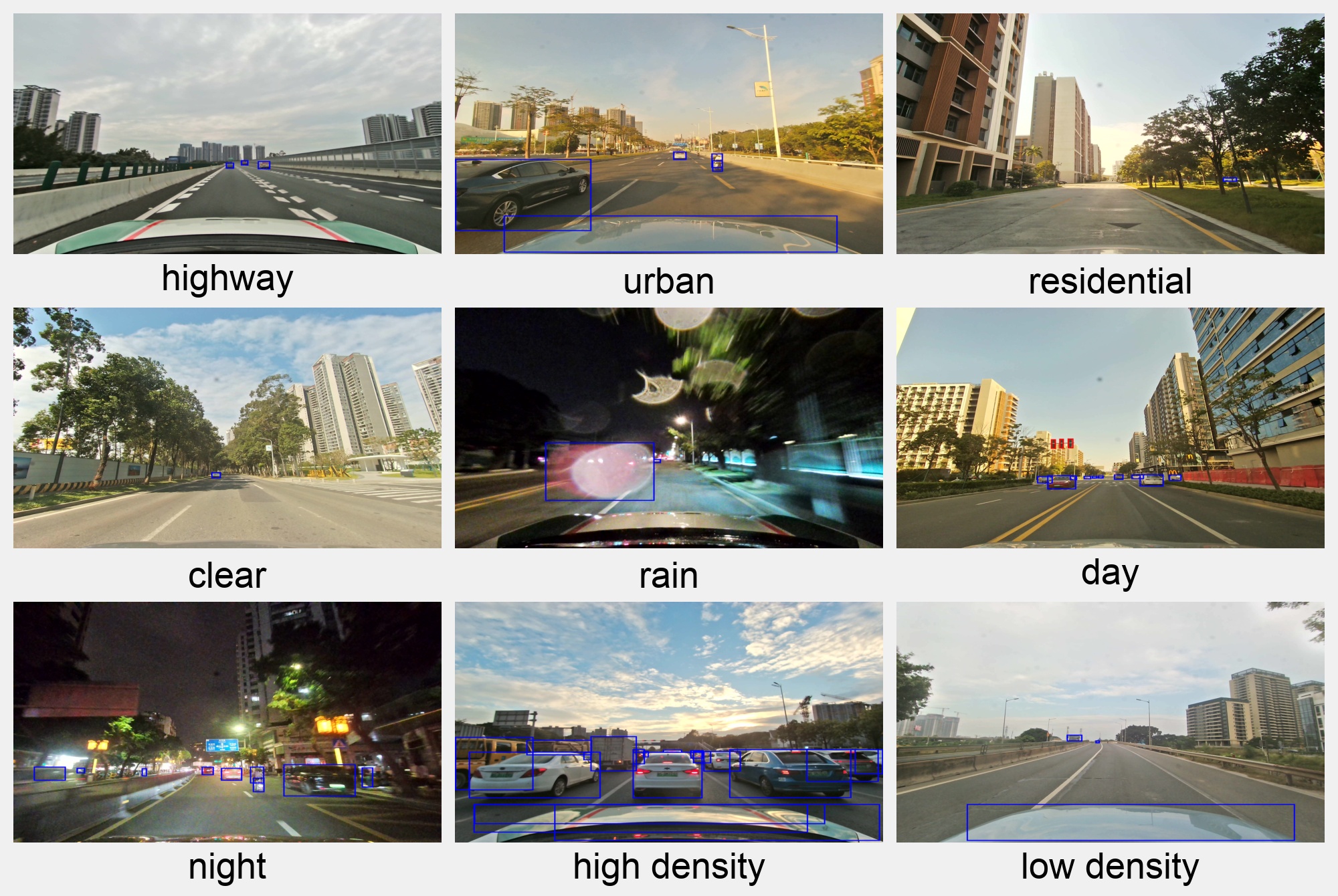}
  \vspace{-0.5em}
  \caption{
      Representative examples of our scenario annotations, covering key conditions such as area type (highway, urban, residential), lighting (day, night), weather (clear, rain), and traffic density (high, low). Each sample is shown with 2D detection boxes to illustrate the annotated traffic participants and objects captured.
    }
  \label{fig:scenario_categories}
\end{figure}

In addition, each frame is annotated with scenario attributes such as area type (\emph{highway, urban, residential}), lighting (\emph{day, dusk, night}), weather (\emph{clear, rain, snow}), road surface type (\emph{paved, unpaved}), vehicle density (\emph{high, mid, low}), VRU density (\emph{high, mid, low}), presence of traffic lights, and traffic sign category (\emph{warning, prohibition, indication}). All scenario annotations are manually labeled by trained annotators based on both visual imagery and vehicle trajectory information. As illustrated in \cref{fig:scenario_categories}, our annotations span all major environmental and traffic conditions encountered in real-world driving.

\subsection{Database Schema}
\label{sec:db_schema}

To efficiently organize and query multimodal data, all raw and processed records are stored in a relational database built on MySQL, following a normalized schema. The database design enables fast cross-referencing between vehicles, sensors, frames, and trajectories while maintaining temporal and spatial consistency.

The core tables and their relationships are illustrated in Fig.~\ref{fig:db_schema}:

\begin{figure}[t]
  \centering
  \includegraphics[width=\linewidth]{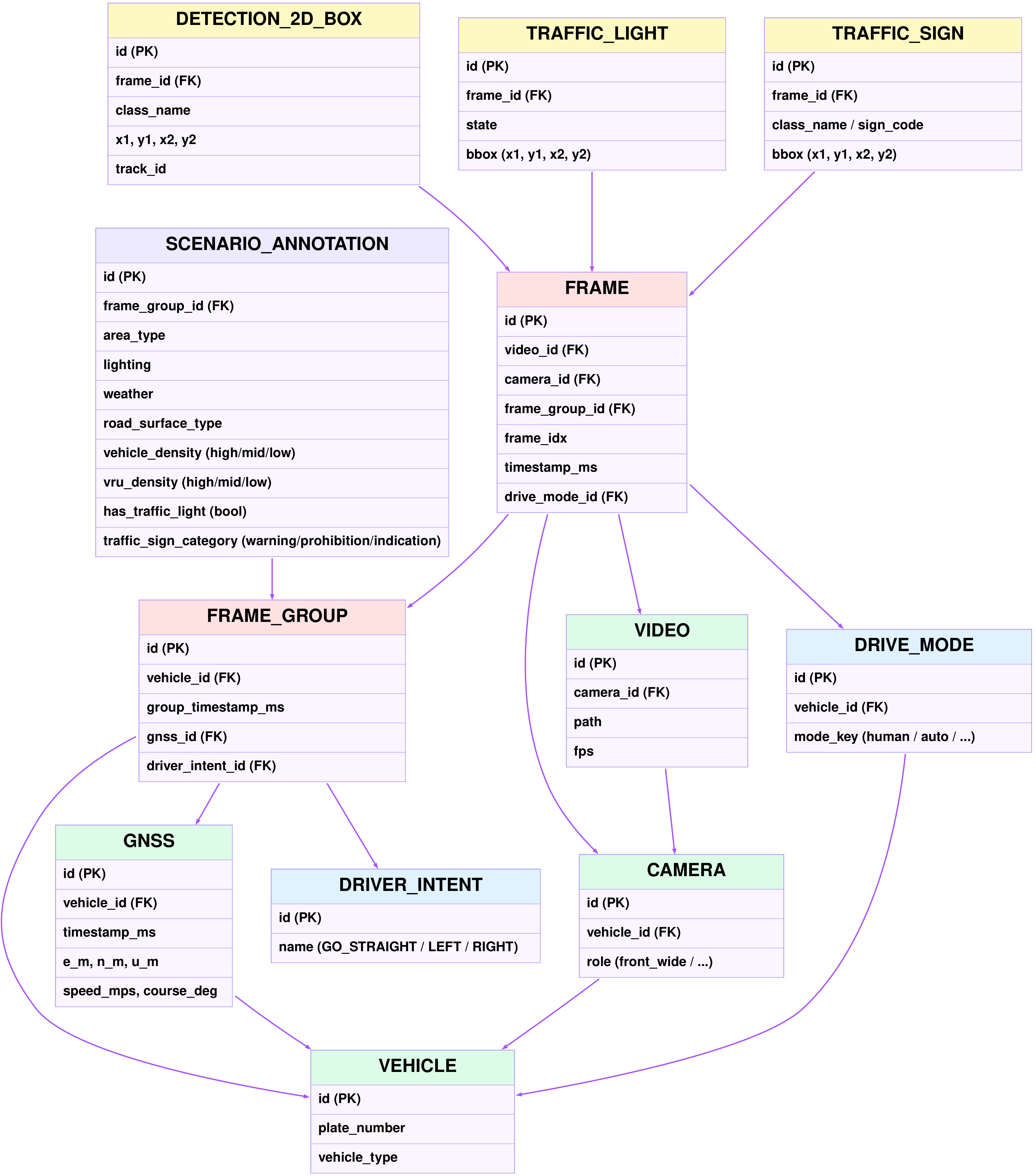}
  \caption{
      Relational schema of the PAVE dataset database, showing core tables and relationships.
  }
  \label{fig:db_schema}
\end{figure}

\begin{itemize}
  \item \textbf{Vehicle}: Stores metadata for each collection vehicle, including plate number, type, and configuration information.
  \item \textbf{Camera}: Defines the mounting position and role of each camera (\eg front\_wide, left\_wide, right\_wide) associated with a vehicle.
  \item \textbf{Video}: Records continuous video segments from each camera, including file path, resolution, frame rate, and duration.
  \item \textbf{FrameGroup}: Groups temporally synchronized multi-camera frames captured at the same timestamp, ensuring precise alignment with GNSS/IMU data.
  \item \textbf{Frame}: Represents an individual frame within a video sequence, each linked to its source video, camera, frame group, and driving mode.
  \item \textbf{GNSS}: Contains timestamped positioning and motion data for each vehicle, including coordinates (E, N, U), velocity, and heading angle.
  \item \textbf{DriveMode}: Indicates the driving state at each frame, such as human or autonomous.
  \item \textbf{DriverIntent}: Labels the high-level driving intention for each frame group, such as going straight, turning left, or turning right.
  \item \textbf{Detection2DBox}: Stores 2D bounding boxes for detected objects, including pedestrians, surrounding vehicles, motorcycles, traffic lights, and traffic signs, each with class name, coordinates, and tracking ID.
  \item \textbf{TrafficLight} and \textbf{TrafficSign}: Record the state of each traffic light and the type of each traffic sign for every frame.
  \item \textbf{ScenarioAnnotation}: Describes scenario-level attributes such as area type, lighting, weather, road surface type, vehicle density, VRU density, presence of traffic lights, and traffic sign category.
\end{itemize}

\subsection{Dataset Analysis}
\label{sec:analysis}

This section presents a comprehensive analysis of our multi-vehicle and multi-model autonomous driving dataset, highlighting its diversity in vehicle models, driving modes, and traffic scenarios. The PAVE dataset integrates several production AVs equipped with different autonomous driving systems, providing a rich foundation for perception, planning, and analysis of AV behaviors.

\paragraph{Model Diversity.}
The PAVE dataset integrates recordings from multiple production AV models, including Tesla, NIO, AITO, Avatr, and XPeng. To preserve data confidentiality, we anonymize these AV models as \textit{Vehicle A–E}. The statistics of these anonymized models, including total hours, average and maximum speed, and mean acceleration, are summarized in \cref{tab:vehicle_dynamics}.

\begin{table}[t]
  \caption{
  Driving statistics across different vehicle platforms (anonymized as A–E). 
  Each platform represents a distinct production vehicle with an independent autonomous driving system.
  }
  \label{tab:vehicle_dynamics}
  \centering
  \resizebox{\columnwidth}{!}{%
  \begin{tabular}{@{}lcccc@{}}
    \toprule
    Vehicle & Hours & Avg. speed (m/s) & Max speed (m/s) & Avg. accel. (m/s$^2$) \\
    \midrule
    A & 20.0 & 9.3 & 31.0 & 0.33 \\
    B & 22.0 & 8.7 & 29.5 & 0.35 \\
    C & 25.0 & 8.4 & 33.0 & 0.37 \\
    D & 28.0 & 14.2 & 35.4 & 0.41 \\
    E & 45.0 & 8.8 & 32.4 & 0.39 \\
    \midrule
    \textbf{Overall} & \textbf{140.0} & \textbf{9.9} & \textbf{35.4} & \textbf{0.38} \\
    \bottomrule
  \end{tabular}%
  }
\end{table}

\paragraph{Geographical Coverage.}
The PAVE dataset covers approximately 88\,km$^2$ of driving data collected across multiple representative regions, including seven major cities in China and the USA. They cover both the downtown area and surrounding suburban roads, capturing diverse traffic scenes including residential streets, arterial roads, high-speed expressways, dense urban roads, and heavily congested city centers that reflect complex, real-world traffic dynamics. This combination of geographically and infrastructurally distinct regions ensures the PAVE dataset encompasses a wide range of driving behaviors.

\paragraph{Driving-Mode Distribution.} 
The PAVE dataset includes both human and autonomous driving, with each segment identified by its driving mode. In total, it comprises over 140\,hours of driving data, including approximately 40\,hours of manual driving and over 100\,hours of autonomous driving.

\paragraph{Scenario Diversity.}

\begin{figure}[t]
  \centering
  \includegraphics[width=1\linewidth]{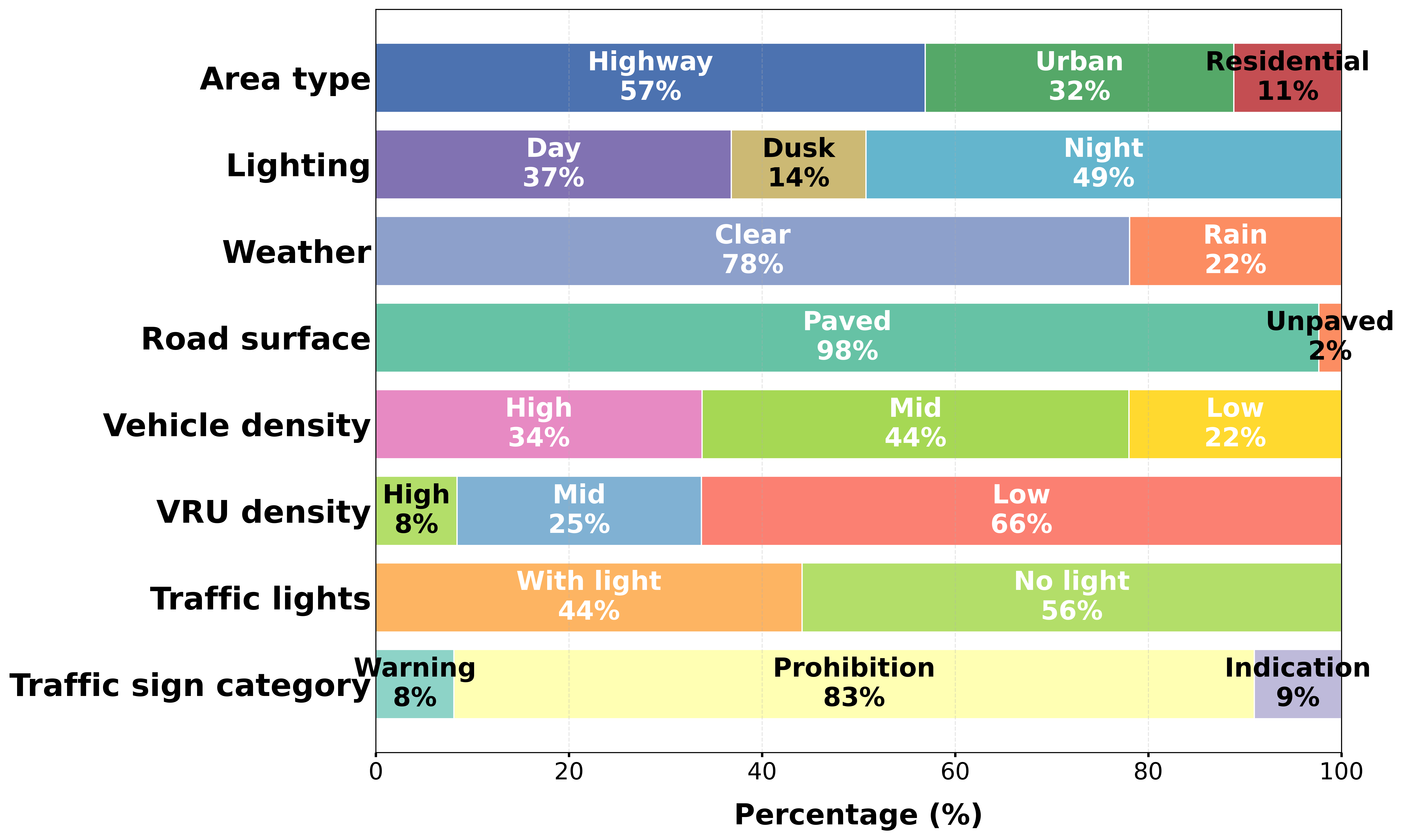}
  \caption{Distribution of scenario types across annotated frames. Vehicle and VRU densities are categorized as \textit{low} (1--5 objects), \textit{medium} (6--15 objects), and \textit{high} ($\geq$16 objects) based on per-frame counts of relevant detections. Traffic sign categories follow the GB~5768 standard~\cite{GB5768.2-2022}: \textit{warning} (yellow triangles), \textit{prohibition} (red circles), and \textit{indication} (blue circles).}
  \label{fig:scenario_gantt}
\end{figure}

The PAVE dataset includes a broad set of scenario types across several key dimensions: \textit{area type} (highway, urban, residential), \textit{lighting} (day, dusk, night), \textit{weather} (clear, rain), \textit{road surface} (paved, unpaved), \textit{vehicle and VRU density}, \textit{traffic-light presence}, and \textit{traffic sign category} following the GB~5768 standard. The detailed distribution of these categories is shown in \cref{fig:scenario_gantt}.

\section{Tasks \& Metrics}
\label{sec:tasks}

The PAVE dataset supports key autonomous driving tasks, including object detection, motion planning, and safety evaluation. These tasks cover the perception, behavior understanding, and safety evaluation of production AVs. In this work, we focus on end-to-end motion planning, which predicts the ego vehicle’s future trajectory from multi-camera images and trajectory history. We also plan to release a safety evaluation framework that will provide quantitative measures of safety and comfort in real-world driving.

\subsection{Detection}
\label{subsec:detection}

The PAVE dataset supports object detection across diverse driving environments. Each frame includes 2D bounding boxes for vehicles, pedestrians, motorcycles, traffic lights, and traffic signs, allowing standard detectors such as Faster R-CNN~\cite{girshick2015fast}, RetinaNet~\cite{lin2017focal}, and YOLOv5~\cite{jocher2020ultralytics} to be trained and evaluated on our data. We also provide full camera intrinsic and extrinsic parameters, enabling 3D extensions using geometric and learning-based methods. Recent approaches such as BEVFormer~\cite{li2024bevformer}, DETR3D~\cite{wang2022detr3d}, and PETR~\cite{liu2022petr} leverage multi-view inputs to predict 3D object positions. Classical methods like ORB-SLAM3~\cite{campos2021orb}, DPT~\cite{ranftl2021vision}, AdaBins~\cite{bhat2021adabins}, and COLMAP~\cite{schonberger2016structure} reconstruct 3D structure or depth from calibrated images. These methods support lifting 2D detections into 3D for downstream tasks.

\paragraph{mean Average Precision (mAP).}
We adopt mAP as the detection metric, following standard practice across object categories and IoU thresholds.
For each class $c$, the Average Precision (AP) is computed as the area under the precision–recall curve:
\begin{equation}
  \mathrm{AP}_c = \int_{0}^{1} p_c(r)\,dr,
  \label{eq:ap}
\end{equation}
The final mean Average Precision is obtained by averaging over all $C$ classes:
\begin{equation}
  \mathrm{mAP} = \frac{1}{C} \sum_{c=1}^{C} \mathrm{AP}_c,
  \label{eq:map}
\end{equation}


\subsection{Motion Planning}
\label{subsec:motion_planning}

The end-to-end motion planning task predicts the ego vehicle’s future trajectory directly from multi-view camera images and its recent trajectory history, without using intermediate perception or map modules. At each timestamp $t$, the model takes synchronized multi-camera images $\mathcal{I}_t$ and the ego vehicle’s past 6 seconds of trajectory sampled at 20 Hz, including position $(x,y)$, velocity $(v_x,v_y)$, acceleration $(a_x,a_y)$, and an intent label $d \in \{\textit{UNKNOWN}, \textit{GO\_STRAIGHT}, \textit{GO\_LEFT}, \textit{GO\_RIGHT}\}$. The model then predicts the future 5 seconds of motion at 20 Hz, producing 100 future waypoints that represent the ego vehicle’s future trajectory.

\subsubsection{Metrics}

We evaluate the motion-prediction performance using displacement-based errors and a safety-related surrogate metric:

\begin{itemize}
  \item \textbf{Average Displacement Error (ADE):}
ADE measures the mean Euclidean distance between the predicted and ground-truth future positions over the prediction horizon:
  \begin{equation}
  \mathrm{ADE} = \frac{1}{H} \sum_{i=1}^{H} 
  \left\| (\hat{x}_i, \hat{y}_i) - (x_i, y_i) \right\|_2 ,
  \label{eq:ade}
  \end{equation}
where $(x_i, y_i)$ and $(\hat{x}_i, \hat{y}_i)$ denote the ground-truth and predicted ego positions at future step $i$, respectively.

  \item \textbf{Final Displacement Error (FDE):}
FDE computes the Euclidean distance between the predicted final waypoint and the ground-truth endpoint:
  \begin{equation}
  \mathrm{FDE} = 
  \left\| (\hat{x}_H, \hat{y}_H) - (x_H, y_H) \right\|_2 .
  \label{eq:fde}
  \end{equation}
   
  
\end{itemize}








\subsection{Safety Evaluation}
\label{subsec:safety}





TTC quantifies potential collision risk between the ego vehicle and a leading vehicle. Following standard practice, TTC is defined only when the ego vehicle is closing in on the leader. At time $t$, let $d_t$ be the longitudinal gap, and let $v_{\text{ego},t}$ and $v_{\text{lead},t}$ be their longitudinal velocities. TTC is computed as: 
\begin{equation}
\mathrm{TTC}_t =
\begin{cases}
\displaystyle 
\frac{d_t}{\,v_{\text{ego},t} - v_{\text{lead},t}\,},
& \text{if } v_{\text{ego},t} > v_{\text{lead},t}, \\[10pt]
+\infty,
& \text{otherwise}.
\end{cases}
\label{eq:ttc}
\end{equation}

where $d_t$ denotes the longitudinal distance between the ego and the preceding vehicle at time \(t\). A smaller TTC value indicates a higher likelihood of unsafe proximity or potential collision. As described in the unified coordinate framework above, all trajectories are represented in the vehicle coordinate system, where the $X$-axis aligns with the forward driving direction. Therefore, the longitudinal velocities $\,v_{\text{ego},t}$ and $v_{\text{lead},t}$ are obtained by projecting their planar velocity vectors onto this vehicle-frame $X$-axis, ensuring consistent TTC computation across all sequences.
\section{Experiments}
\label{sec:experiments}


We evaluate both object detection and motion planning tasks on the PAVE dataset, followed by the performance as a baseline.

\subsection{Object Detection Benchmark}
\label{subsec:det_exp}

We train a YOLO-based~\cite{jocher2020ultralytics} detector on the PAVE dataset to evaluate 2D object detection performance. We evaluate it on a separate validation split and report AP@0.5, AP@0.75, and mAP for each category.

\begin{table}[t]
  \centering
  \scriptsize 
  \caption{
    Per-category 2D detection performance of the YOLO baseline.
  }
  \label{tab:detection_results}
  \begin{tabular}{c|c|c|c|c|c|c|c}
    \toprule
    \textbf{Cat.} & Car & Bus & Truck & Moto & Bicycle & Person & Overall \\
    \midrule
    \textbf{AP@0.5}& 0.471 & 0.313 & 0.396 & 0.365 & 0.373 & 0.426 & \textbf{0.391} \\
    \textbf{AP@0.75}& 0.418 & 0.297 & 0.358 & 0.272 & 0.314 & 0.360 & \textbf{0.337} \\
    \textbf{mAP}    & 0.387 & 0.283 & 0.347 & 0.247 & 0.275 & 0.322 & \textbf{0.310} \\
    \bottomrule
  \end{tabular}
\end{table}

As shown in \cref{tab:detection_results}, the YOLO baseline performs well on larger and more frequently observed objects such as cars, trucks, and persons. Performance on smaller objects like motorcycles and bicycles is lower, mainly because these targets often appear at long ranges or under partial occlusion. These results reflect the realistic distribution and visual complexity of the PAVE dataset.

\subsection{End-to-End Motion Planning Model}

We adopt the E2EDriver~\cite{E2EDriver2025} architecture as our end-to-end motion planning baseline, which directly predicts the future ego trajectory from raw multi-sensor inputs. The model encodes multi-view camera images using a lightweight SqueezeNet combined with a Vision Transformer (ViT) to extract both local and global scene features, while the ego vehicle’s trajectory history and driving intent are encoded by multilayer perceptrons (MLPs). These representations are fused into a unified latent feature, and the decoder outputs a fixed-horizon trajectory represented by $(x, y)$ waypoints. 



\subsection{Results}

We evaluate the end-to-end motion planning baseline separately under two independent settings: driving mode and area type. For each setting, we report the ADE and FDE to show the motion task results.

\begin{table*}[t]
  \centering
  \caption{
    End-to-end motion planning baseline performance table.
    ADE and FDE are reported in meters.
    Abbreviations: Hum.=Human, Aut.=Autonomous, Hwy.=Highway, Urb.=Urban,
    Day=Daytime, Ngt.=Night, Clr.=Clear,
    L/M/H=Low/Medium/High density, W/N=With/No traffic light,
    Full=Full Dataset.
  }
  \label{tab:planning_results}
  \begin{tabular}{l|
    cc|cc|cc|cc|ccc|cc|c}
    \toprule
    \textbf{Category} &
    \multicolumn{2}{c|}{Driving Mode} &
    \multicolumn{2}{c|}{Area Type} &
    \multicolumn{2}{c|}{Lighting} &
    \multicolumn{2}{c|}{Weather} &
    \multicolumn{3}{c|}{Vehicle Density} &
    \multicolumn{2}{c|}{Traffic Light} &
    Overall \\
    \midrule
    \textbf{Setting} &
    Hum. & Aut. &
    Hwy. & Urb. &
    Day & Ngt. &
    Clr. & Rain &
    L & M & H &
    W & N &
    Full \\
    \midrule
    \textbf{ADE (m)} &
    1.20 & 1.76 &
    1.76 & 1.27 &
    1.49 & 1.62 &
    1.49 & 1.76 &
    1.66 & 1.53 & 1.55 &
    1.67 & 1.40 &
    \textbf{1.47} \\
    \midrule
    \textbf{FDE (m)} &
    4.76 & 10.40 &
    11.38 & 4.59 &
    7.86 & 8.43 &
    8.39 & 7.59 &
    8.87 & 8.73 & 7.58 &
    7.94 & 8.56 &
    \textbf{8.16} \\
    \bottomrule
  \end{tabular}
\end{table*}




As reported in \cref{tab:planning_results}, the model achieves lower ADE and FDE on human-driving data than on autonomous-driving data, indicating that human trajectories are easier to predict. The model also performs better in urban areas than on highways, since higher speeds on highways lead to larger errors. Daytime scenes give slightly lower errors than nighttime, and clear and rainy weather show similar results. Vehicle density has only a small impact. Overall, the model trained on the full dataset reaches an ADE of 1.47 and an FDE of 8.16.

\begin{figure*}[t]
    \centering
    \includegraphics[width=\textwidth]{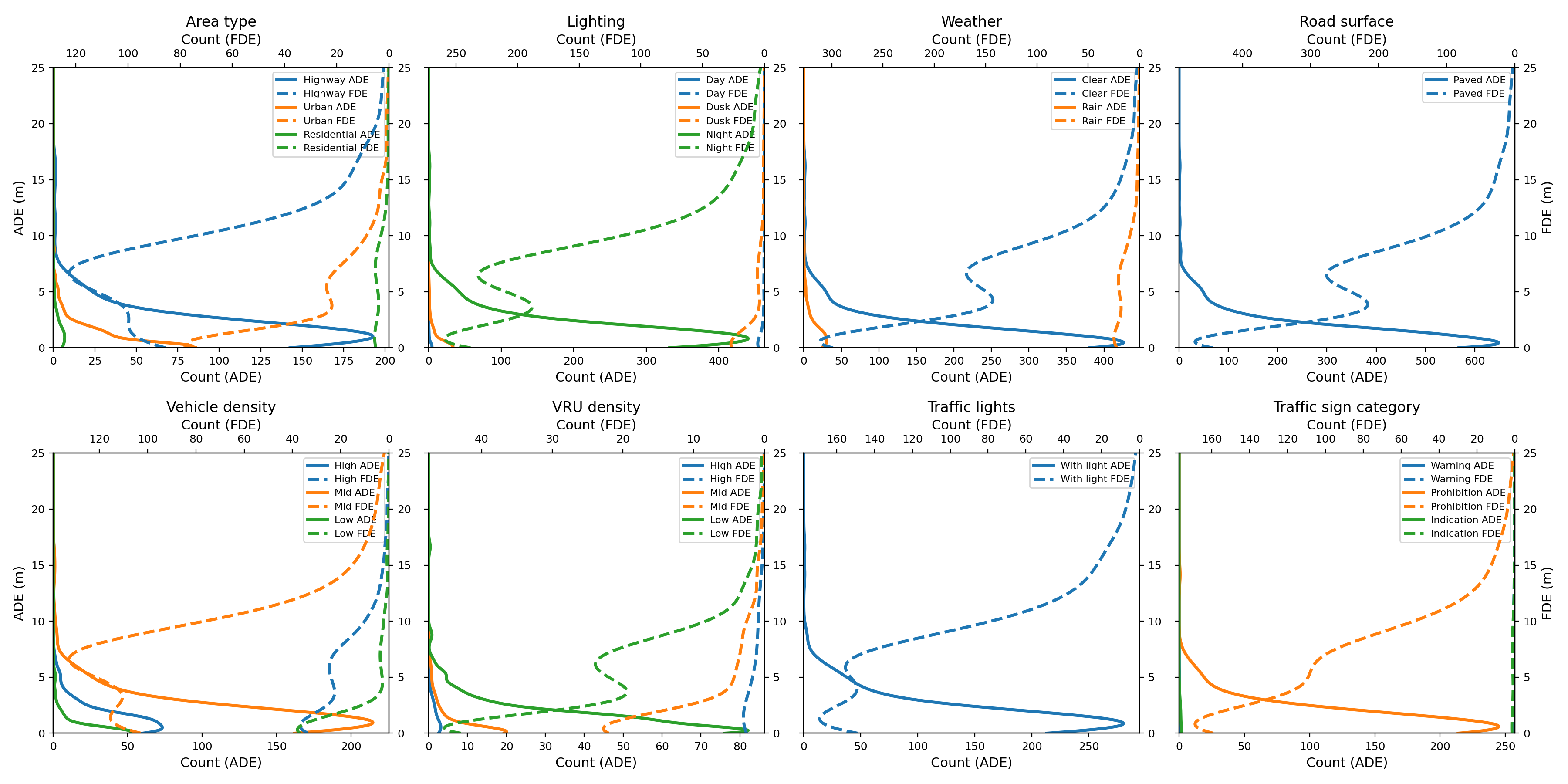}
    \caption{
        Distributions of ADE and FDE errors across different traffic scenarios. Each subplot corresponds to one scenario dimension (area type, lighting, weather, road surface, vehicle/VRU density, traffic lights, and traffic sign categories), with solid lines indicating ADE and dashed lines indicating FDE.
    }
    \label{fig:ade_fde_scenarios}
\end{figure*}

Figure~\ref{fig:ade_fde_scenarios} shows the ADE and FDE distributions under different scenario attributes. Across all scenario dimensions, a consistent trend can be observed: as the driving environment becomes more complex, the prediction uncertainty increases, resulting in larger ADE and FDE values. Simpler conditions lead to more concentrated and lower-error distributions, while more challenging situations introduce heavier long tails.

\subsection{Discussion}
The current results reflect the behavior of the model under the subset of scenarios included in this release. While the model performs consistently across most conditions, its predictions vary across driving modes, speed ranges, and lighting conditions. As the PAVE dataset continues to grow with more autonomous-driving segments, nighttime sequences, adverse weather, and high-density traffic, future evaluations will provide more reliable insights.

\section{Conclusion}
\label{sec:conclusion}

We introduced the PAVE dataset, a large-scale real-world autonomous driving dataset containing synchronized multi-camera imagery, GNSS/IMU measurements, and explicit driving-mode annotations. The dataset covers both human and autonomous driving and spans diverse geographic areas, lighting conditions, weather types, and traffic densities. These characteristics make it suitable for studying perception, motion prediction, and end-to-end motion planning under realistic driving conditions.

Each sequence provides wide-view camera inputs and accurate trajectory data in a unified coordinate framework, enabling fine-grained analysis of vehicle behavior across different modes and environments. By integrating sensor streams and annotations, the dataset offers a consistent platform for evaluating how well models generalize to complex, real-world scenarios.

We further benchmarked an adapted end-to-end planning model on the dataset. The model achieves an ADE of 1.47\,m and an FDE of 8.16\,m, demonstrating stable performance across a broad set of conditions. These results provide an initial reference for future research and illustrate the value of the dataset for trajectory planning evaluation.

The current release reflects only a subset of all planned driving routes, vehicle platforms, and environmental conditions. Some scenario types—such as nighttime autonomous-driving segments, adverse weather, and dense urban traffic—are still limited in scale. As the dataset continues to expand, we expect future versions to provide broader coverage and more statistically reliable evaluation results. 

{
    \small
    \bibliographystyle{ieeenat_fullname}
    \bibliography{main}
}


\end{document}